\pdfoutput=1

\documentclass[11pt]{article}

\usepackage[]{acl}

\usepackage{times}
\usepackage{latexsym}

\usepackage[T1]{fontenc}
\usepackage{amsfonts}

\usepackage[utf8]{inputenc}

\usepackage{microtype}

\usepackage{amsmath}
\usepackage{graphicx}

\usepackage{float}
\usepackage{pgfplots}
\usepackage{subfigure}
\usepackage{booktabs}
\usepackage{multicol}
\usepackage{multirow}
\usepackage{arydshln} 
\usepackage{CJKutf8}

\usepackage{bbding}

\title{Unsupervised Full Constituency Parsing with \\Neighboring Distribution Divergence}

\author{Letian Peng$^{1,2,3,\dag}$, Zuchao Li$^{1,2,3,\dag}$, and Hai Zhao$^{1,2,3}$\thanks{$\ $  Corresponding author. $^\dag$ These authors made equal contribution. This work was supported by Key Projects of National Natural Science Foundation of China (U1836222 and
61733011).}\\
$^{1}$Department of Computer Science and Engineering, Shanghai Jiao Tong University \\
	$^{2}$Key Laboratory of Shanghai Education Commission for Intelligent Interaction \\ and Cognitive Engineering, Shanghai Jiao Tong University, Shanghai, China\\
	$^{3}$MoE Key Lab of Artificial Intelligence, AI Institute, Shanghai Jiao Tong University \\
  {\tt \small \{zxc-00,charlee\}@sjtu.edu.cn, zhaohai@cs.sjtu.edu.cn}}
  
\begin{document}
\maketitle
\begin{abstract}
Unsupervised constituency parsing has been explored much but is still far from being solved. Conventional unsupervised constituency parser is only able to capture the unlabeled structure of sentences. Towards unsupervised full constituency parsing, we propose an unsupervised and training-free labeling procedure by exploiting the property of a recently introduced metric, Neighboring Distribution Divergence (NDD), which evaluates semantic similarity between sentences before and after editions. For implementation, we develop NDD into Dual POS-NDD (DP-NDD) and build "molds" to detect constituents and their labels in sentences. We show that DP-NDD not only labels constituents precisely but also inducts more accurate unlabeled constituency trees than all previous unsupervised methods with simpler rules. With two frameworks for labeled constituency trees inference, we set both the new state-of-the-art for unlabeled F1 and strong baselines for labeled F1. In contrast with the conventional predicting-and-evaluating scenario, our method acts as an plausible example to inversely apply evaluating metrics for prediction.
\end{abstract}

\section{Introduction}
Constituency parsing is a basic but crucial parsing task in natural language processing. Constituency parsers are required to build parsing trees for sentences which consist of spans representing constituents such as noun phrase and verb phrase. Parsed constituency trees can be applied to many downstream systems \cite{DBLP:journals/coling/LeeCPCSJ13,chen-etal-2015-shallow,DBLP:journals/cogcom/ZhongCH20}.

Since the introduction of deep learning into natural language processing, supervised neural networks have achieved remarkable success in constituency parsing \cite{DBLP:conf/acl/KleinK18,DBLP:conf/aaai/LiuZS18,DBLP:conf/acl/NguyenNJL20,DBLP:conf/ijcai/ZhangZL20}. Unfortunately, the need of large annotated datasets limits the performance of supervised systems on languages of low resources. As the result, many unsupervised systems have been proposed for constituency parsing \cite{DBLP:conf/naacl/DrozdovVYIM19,DBLP:conf/iclr/KimCEL20,DBLP:conf/acl/ShenTZBMC20,DBLP:conf/acl/SahayNMRI21} by exploiting unlabeled corpus.

\begin{figure}
\centering
\includegraphics[width=0.5\textwidth]{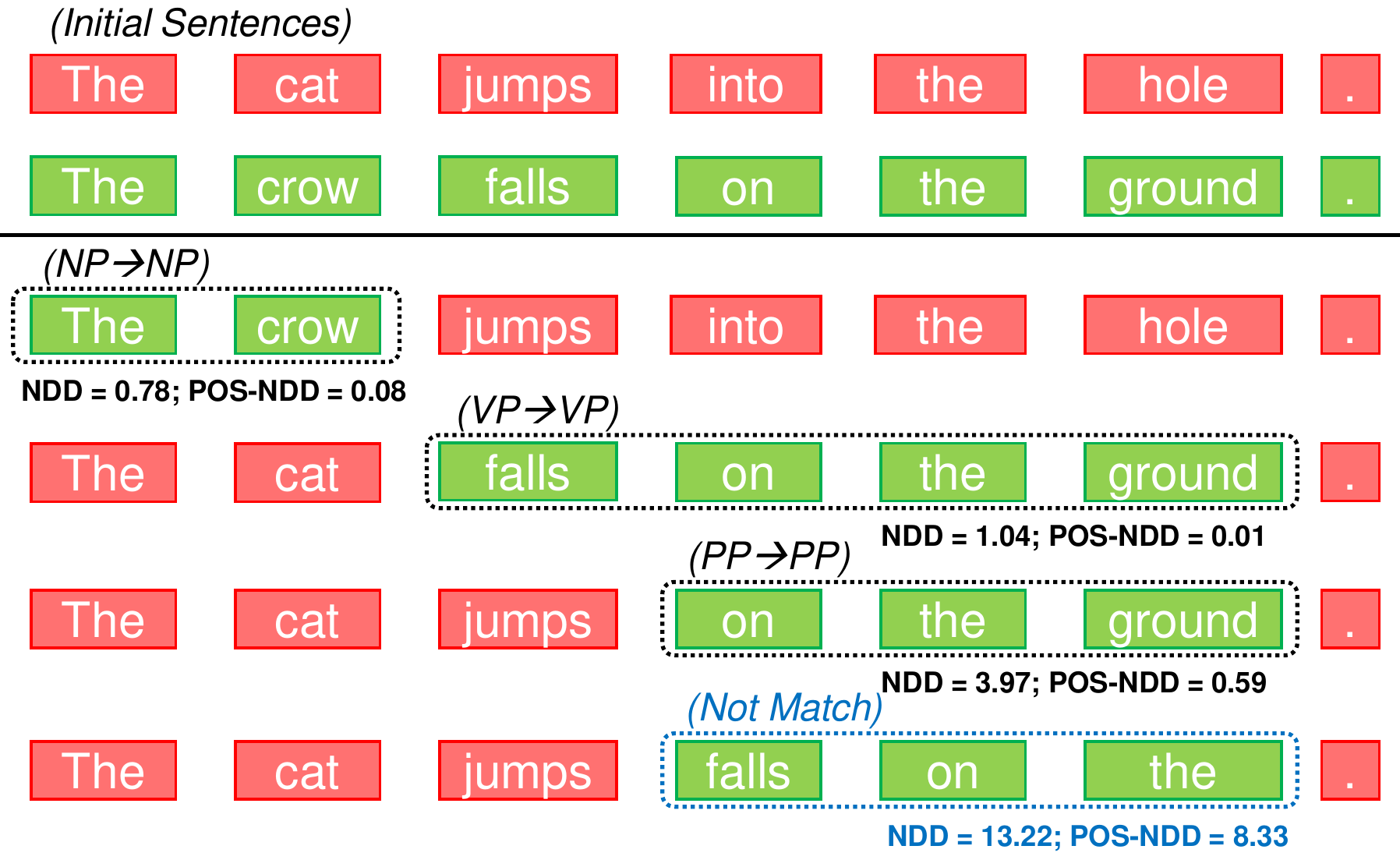}
\caption{Examples for substitution property of constituents among sentences. Neighboring Distribution Divergence performs well on detecting plausible substitutions.}
\label{fig:example}
\end{figure}

\begin{figure}
\centering
\includegraphics[width=0.5\textwidth]{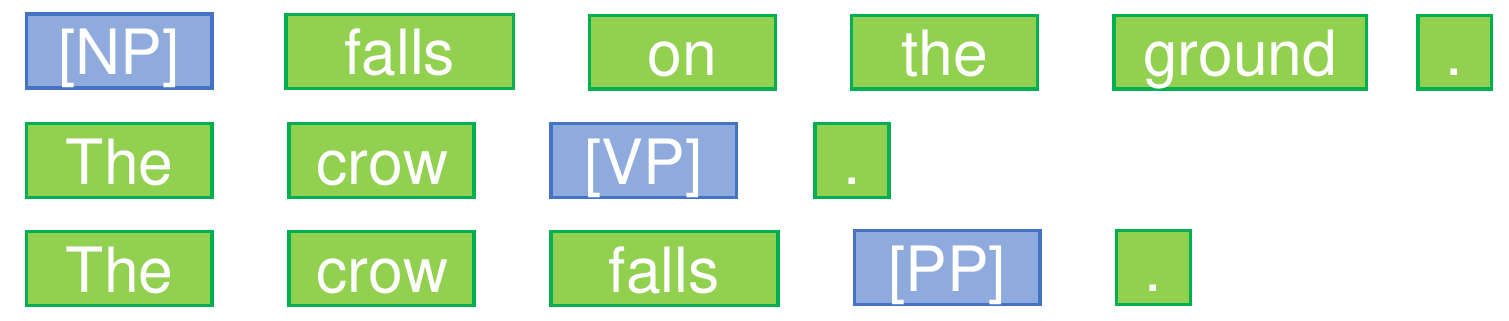}
\caption{Molds constructed from examples in Figure~\ref{fig:example}.}
\label{fig:mold}
\end{figure}

\begin{figure*}
\centering
\includegraphics[width=0.99\textwidth]{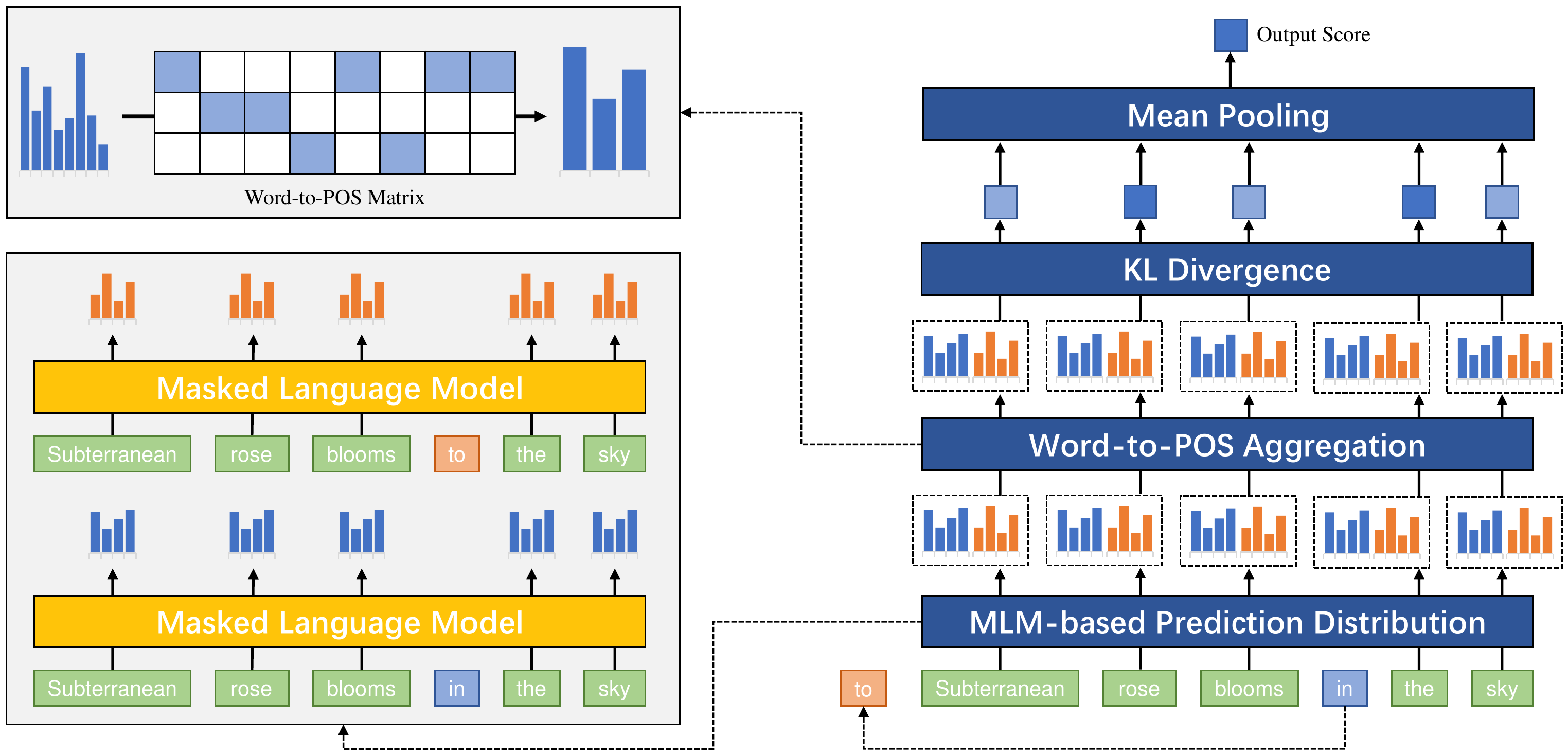}
\caption{Calculating procedure for POS-NDD.}
\label{fig:ndd}
\end{figure*}

However, current unsupervised constituency parsing systems are still far from a real full procedure, especially for the reason that most of these systems only induct an unlabeled structure of the constituency tree. Towards the unsupervised full constituency parsing, we exploit a recently proposed metric, Neighboring Distribution Divergence (NDD) \cite{DBLP:journals/corr/abs-2110-01176}, to automatically detect and label constituents in sentences. NDD is a Pre-trained Language Model-based (PLM-based) \cite{DBLP:conf/naacl/DevlinCLT19} metric and detects semantic changes caused by editions. We use NDD in a tricky way to adapt it to constituency parsing.

We first put forward the substitution property of constituents. As shown in Figure~\ref{fig:example}, if we substitute a constituent with another constituent with the same label, the edited sentence will still be plausible in syntax and semantics. Therefore, if the substitution by a span to an annotated constituent results in a plausible sentence, that span will be of high probability to be a constituent with the same label. Thus, metrics like perplexity may be able to detect successful substitutions. However, according to \cite{DBLP:journals/corr/abs-2110-01176}, perplexity lacks the capacity to provide reasonable comparison between sentences and has been prominently outperformed by NDD on most related tasks. So for the detecting metric, we instead use NDD which is able to detect very precise semantics similarity.

In practice, we construct very few detectors called "molds" as shown in Figure~\ref{fig:mold}. We judge whether a span to be a constituent by fill it into the phrase mask (\textit{[NP]}, \textit{[VP]}, ...) and use NDD to detect the semantic similarity between the filled sentence and the initial sentence. If NDD is under a certain threshold, the span will be predicted to be a constituent.

To further boost the efficiency and performance of our method, we modify NDD into POS-NDD which only considers the likeliness of POS sequences since the initial NDD is too sensitive to precise semantic difference. Also, we use a dual detecting method which evaluates both the \textbf{substitution to mold} and \textbf{substitution from mold} to better link constituents with the same label together. We named our final metric Dual POS-NDD (DP-NDD).

We experiment on Penn Treebanks to construct labeled constituency trees and label predicted treebanks from other unsupervised constituency parsers. Results from our experiments verify DP-NDD to be capable of inducting labeled constituency trees and labeling unlabeled constituents. Based on DP-NDD molds, we introduce two novel frameworks for unsupervised constituency parsing. Our algorithm parses by following simple rules but results in remarkable results which outperform all previous unsupervised parsers on WSJ test dataset. Our algorithms set the first strong baseline in recent years for labeled F1 score. Our main contributions are concluded as follows:

\begin{itemize}
\item We propose an unsupervised method for full constituency parsing which involves constituent labeling.
\item We introduce two novel frameworks for unsupervised constituency parsing, which set a new state-of-the-art for unlabeled F1 and strong baselines for labeled F1.
\item We introduce variants of NDD, POS-NDD and DP-NDD which are less sensitive to semantic difference between sentences and perform well for constituent detecting.
\item We make a try for modeling parsing in an editing form, which is different from the conventional practice which aids editions with parsing results.
\end{itemize}

\section{Neighboring Distribution Divergence}
\subsection{Background}

We give a brief description for the NDD metric in this section as the background for further discussion. More details like motivation and explanation can be referred to \cite{DBLP:journals/corr/abs-2110-01176}. 

Given an $n$-word sentence $W = [w_1, w_2, \cdots, w_n]$, we use an edition $E$ to convert $W$ to an edited sentence $W' = E(W)$. As we only use substitution for unsupervised constituency parsing, we limit $E$ to a substituting operation which substitutes $i$-th to $j$-th word in $W$ with a span $V = [v_1, v_2, \cdots, v_m]$. 

\begin{equation*}
\centering
\small
\begin{aligned}
W' = E(W) = [w_1, \cdots, w_{i-1}, v_1, \cdots, v_m, w_{j+1}, \cdots, w_n]
\end{aligned}
\end{equation*}

Then we evaluate the semantic disturbance on the overlapped part $[w_1, \cdots, w_{i-1}, w_{j+1}, \cdots, w_n]$ between the initial and edited sentences. For estimation, we use masked language model to get the distribution of predicted words for each masked position before and after the edition.

\begin{equation*}
\centering
\small
\begin{aligned}
W^{mask}_i &= [w_1, \cdots, w_{i-1}, \textrm{[MASK]}, w_{i+1}, \cdots, w_n];\\
R &= PLM(W^{mask}_i); d_i = \textrm{softmax}(R_i) \in \mathbb{R}^{c}
\end{aligned}
\end{equation*}

We first mask the $i$-th word in $W$ and use the PLM to predict the distribution $d_i$ on the masked position. Here, $d_i$ is a $\mathbb{R}^{c}$ tensor which refers to the existence probability of the words in a $c$-word dictionary of the PLM. We do this for the overlapped part mentioned above both in $W$ and $W'$.

After we get the predicted distributions for $W$ and $W'$, we use KL divergence to calculate the difference between the two distributions.

\begin{equation*}
\centering
\small
\begin{aligned}
div_i = D_{KL}(d'_i||d_i) = \sum_{j=1}^c d'_{ij}\log(\frac{d'_{ij}}{d_{ij}})
\end{aligned}
\end{equation*}

Finally, we integrate the divergence values together via a mean pooling layer.

\begin{equation*}
\centering
\small
\begin{aligned}
\textrm{NDD}(W, W') &= \sum_{k \in [1, \cdots, i-1, j+1, \cdots, n]} \frac{{div}_k}{n-(j-i+1)} \\
\end{aligned}
\end{equation*}

According to the cases in \cite{DBLP:journals/corr/abs-2110-01176}, NDD is capable of capturing precise semantics changes. We will show in Section~\ref{2.3} how to use modified NDD to construct molds for unsupervised constituency parsing.

\subsection{POS-NDD}

\begin{table}
\centering
\small
\scalebox{0.75}{ 
\begin{tabular}{p{5cm}cccc}
\toprule
Sentence & Sem. & Str. & POS-NDD & NDD\\
\midrule
The spider built its nest in the cave. & - & - & 0.00 & 0.00\\
\midrule
The spider \underline{made} its nest in the cave. & \XSolidBrush & \XSolidBrush & 0.69 & 2.67\\
The spider \underline{caught the pests} in the cave. & \Checkmark & \XSolidBrush & 0.81 & 7.45\\
The spider \underline{a wasted bridge} in the cave. & \Checkmark & \Checkmark & 6.42 & 18.39\\
\bottomrule
\end{tabular}
}
\caption{Comparison between NDD and POS-NDD for semantic and structural change detection. The initial sentence is \textit{"The spider built its nest in the cave."} \textbf{Sem.}: If there is a semantic change. \textbf{Str.}: If there is a structural change. } 
\label{tab:pndd}
\end{table}

NDD performs well on supervising editions, but it might be too sensitive to some precise semantic difference. To adapt NDD to constituency parsing, we modify NDD's calculating procedure to make it concentrate on the structural rather than semantic difference. 

To do so, we gather the predicted words with the same POS together by summing up the existence probability of them. For implementation, we construct a word-to-POS matrix $M$ as shown in Figure~\ref{fig:ndd}. $M$ is a 2-dimension tensor of shape $\mathbb{R}^{p\times c}$ where $p$ is the number of POS classes and $c$ is the scale of PLM's dictionary. $M$ is constructed following the rule as follows:

\begin{equation*}
\centering
\small
M_{ij} = \left\{
\begin{aligned}
0, & \textrm{ if $j$-th word dictionary not in $i$-th POS class} \\ 
1, & \textrm{ if $j$-th word dictionary in $i$-th POS class} 
\end{aligned}
\right.
\end{equation*}

With $M$, we gather the existence probability of words in the same POS class together and calculate the KL divergence for POS-NDD. The weighted sum in POS-NDD calculation is the same as in NDD

\begin{equation*}
\centering
\small
\begin{aligned}
q_i &= Md_i, q'_i = Md'_i\\
div^{pos}_i &= D_{KL}(q'_i||q_i) = \sum_{j=1}^p q'_{ij}\log(\frac{q'_{ij}}{q_{ij}})
\end{aligned}
\end{equation*}

The comparison between the initial NDD and modified POS-NDD is presented in Table~\ref{tab:pndd}. In the first example, we edit the sentence while keeping both the semantics and structure unchanged. The edition results in rather low values for both NDD and POS-NDD. In the second example, our edition does not convert the sentence's structure but result in difference semantics. Initial NDD is sensitive to this change as its value raises to nearly $\times 3$. In contrast, POS-NDD is less likely to be affected by semantics and concentrates more on the sentence structure The last example includes an edition which breaks the sentence's structure as it substitutes a verb phrase by a probably noun phrase. As the value of POS-NDD raises to almost $\times 8$, POS-NDD is verified to detect this anomaly.

\subsection{NDD-based Dual Mold}
\label{2.3}

\begin{figure}
\centering
\includegraphics[width=0.5\textwidth]{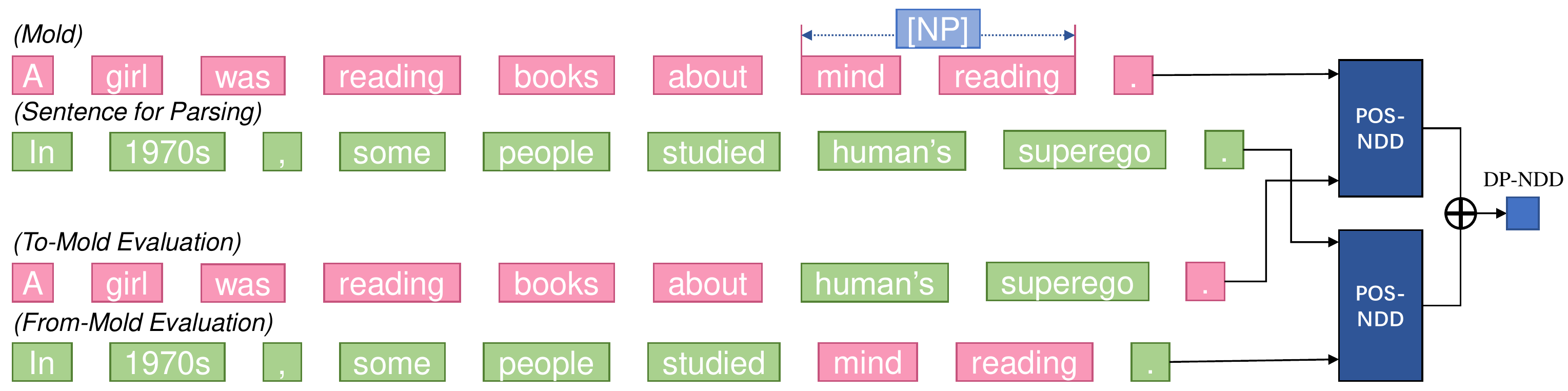}
\caption{Dual mold for detecting constituents.}
\label{fig:dual}
\end{figure}

Based on POS-NDD, we build "molds" which are able to discern constituents in sentences. Our mold is defines as a quaternion $(W, i, j, l)$ where $W$ is an $n$-word sentence. $i, j$ refers to the start and end position of the span for substitution. $l$ refers to the constituent's label. If we want to evaluate the probability of a span $V[s:t]$ (words from $s$-th to $t$-th position in an $m$-word sentence $V$) to a constituent with the label $l$, we will substitute $W[i:j]$ with $V[s:t]$ and calculate the POS-NDD between the sentence before and after the substitution. 

\begin{equation*}
\centering
\small
\begin{aligned}
S^{l,tm}_{s,t} &= \textrm{POS-NDD}(W, W')\\
W' &= [w_1, \cdots, w_{i-1}, v_{s}, \cdots, v_{t}, w_{j+1}, \cdots, w_n]
\end{aligned}
\end{equation*}

We call this score \textbf{To-Mold} score as it is obtained by substituting spans in molds. Likewise, we also have a \textbf{From-Mold} score which is obtained by substituting spans in sentences for parsing with spans in molds. 

\begin{equation*}
\centering
\small
\begin{aligned}
S^{l,fm}_{s,t} &= \textrm{POS-NDD}(V, V')\\
V' &= [v_1, \cdots, v_{s-1}, w_{i}, \cdots, w_{j}, v_{t+1}, \cdots, v_m]
\end{aligned}
\end{equation*}

We finally add To-Mold and From-Mold scores together for whole evaluation,

\begin{equation*}
\centering
\small
\begin{aligned}
S^{l}_{s,t} = S^{l,tm}_{s,t} + S^{l,fm}_{s,t}
\end{aligned}
\end{equation*}

which forms a dual calculating procedure as shown in Figure~\ref{fig:dual}. We thus name our method Dual POS-NDD. A lower DP-NDD score $S$ refers to less disturbance in substituting to and by a constituent with label $l$ and will thus reflects the likelihood of the span to be a constituent with the same label. 

\section{Constituency Tree Constructing}
In this section, we will introduce two frameworks that we use in experiments to generate labeled constituency trees. 

\subsection{Labeled Span Generating}

Labeled Span Generating (LSG) is to directly generate labeled spans with DP-NDD molds and then integrate spans with different labels together to construct the full labeled tree. Our LSG algorithm is much simpler than previous rules-based systems as it only requires $4$ steps for constituency parsing.

\begin{itemize}
\item \textbf{Candidate Selection} We first use simple linguistic rules to sample some candidates for a constituent label. For a span $V[s:t]$, we match the POS tags of $V[s-1]$, $V[s]$, $V[t]$, $V[t+1]$ to a POS list to roughly decide whether the span is a plausible candidate for constituent or not. 
\item \textbf{DP-NDD Scoring} We then use our Dual POS-NDD molds to score the sampled candidates as previously described. For some labels, there are multiple molds for evaluation as difference exists among constituents with the same label. We choose the minimal value of DP-NDD scores from the molds.
\item \textbf{Conflict Removing} After scoring, we remove spans which conflict with previously parsed span. Conflicting spans are those overlapping with previous spans by $(s < s', s' < t, t < t')$ or $(s' < s, s < t', t' < t)$.
\item \textbf{Filtering and Overlapping Removing} Finally, we filter the spans by only keeping the spans with DP-NDD scores under a certain threshold. Then, we remove spans overlapped with other spans of the same label. If $(s < s', s' < t, t < t')$ or $(s' < s, s < t', t' < t)$, we only keep the span with higher DP-NDD. But if $(s < s', t' < t)$ or $(s' < s, t < t')$, we add a tolerance factor to the algorithm to keep both spans if the difference between the two scores is lower than the tolerance. 
\end{itemize}

We execute the $4$ steps above for each label and finally integrate spans parsed from each iteration together to construct the whole labeled constituency tree. 

\subsection{Unlabeled Tree Labeling}

Unlabeled Tree Labeling (ULT) is to first use a parsing algorithm to induct unlabeled treebanks from sentences, and then use DP-NDD molds to label the spans in the tree. Our UTL only annotates the edges in the tree with no changing in the tree structure. For each label, we use a mold to calculate the DP-NDD score. The span is labeled as label of the mold to minimize the DP-NDD. In practice, we maximize the exponential of negative DP-NDD. 

\begin{equation*}
\centering
\small
\begin{aligned}
l_{s,t} = \mathop{\textrm{argmax}}\limits_l(e^{-S^l_{s,t}})
\end{aligned}
\end{equation*}

We further refine the prediction by incorporating POS tags, we use the posterior probability collected before for approximation to induct the label of a span using the POS of start and end words. 

\begin{equation*}
\centering
\small
\begin{aligned}
l_{s,t} &= \mathop{\textrm{argmax}}\limits_l(\alpha e^{-S^l_{s,t}})\\
\alpha &= p(l|\textrm{POS}(V[s]))p(l|\textrm{POS}(V[t]))
\end{aligned}
\end{equation*}

where we add $\alpha$ as a modifier to incorporate POS-based probability into prediction.

\section{Experiment}
\subsection{Data and Configuration}

We experiment our parsing algorithm on Penn Treebank for Constituency Parsing. As our method is training-free, we only use the first 50 sentences in the development dataset to construct molds and we handcraft some simple molds. We do not use the training dataset and test our algorithm on the test dataset. We use molds of a number fewer than 25. We apply BERT-base-cased \cite{DBLP:conf/naacl/DevlinCLT19} as the PLM for calculating DP-NDD. We also have two configurations for thresholds and tolerances in LSG. A strict configuration will produce fewer predicted spans and will thus result in higher labeled F1 scores while a loose configuration will in opposite result in higher unlabeled F1 scores. For ULT, we use DIORA+PP (Post-processing) \cite{DBLP:conf/naacl/DrozdovVYIM19} which is a strong baseline for unsupervised constituency parsing to induct the unlabeled treebanks. For probability approximation in POS-based refinement for UTL, we only use POS tags in development dataset. Specific molds, POS-based rules, thresholds and tolerances can be referred to Appendix~\ref{sec:config}.

\subsection{Main Result}

\begin{table}
\centering
\small
\scalebox{1.0}{ 
\begin{tabular}{lcc}
\toprule
Method & UF1 & LF1$^*$\\
\midrule
LB&13.1&-\\
RB&16.5&-\\
RL-SPINN \cite{DBLP:conf/aaai/ChoiYL18}&13.2&-\\
ST-Gumbel - GRU \cite{DBLP:conf/iclr/YogatamaBDGL17}&22.8&-\\
\midrule
PRPN \cite{DBLP:conf/iclr/ShenLHC18}&38.3&-\\
BERT-base \cite{DBLP:conf/iclr/KimCEL20}&42.3&-\\
ON-LSTM \cite{DBLP:conf/iclr/ShenTSC19}&47.7&-\\
XLNet-base \cite{DBLP:conf/iclr/KimCEL20}&48.3&-\\
DIORA \cite{DBLP:conf/naacl/DrozdovVYIM19}&48.9&-\\
Tree-T \cite{DBLP:conf/emnlp/WangLC19}&49.5&-\\
StrctFormer \cite{DBLP:conf/acl/ShenTZBMC20}&54.0&-\\
\midrule
PRPN+PP&45.2&-\\
DIORA+PP&55.7&-\\
DIORA+PP+Aug \cite{DBLP:conf/acl/SahayNMRI21}&58.3&-\\
\midrule
Neural PCFG \cite{DBLP:conf/acl/KimDR19}&50.8&-\\
Compound PCFG \cite{DBLP:conf/acl/KimDR19}&55.2&-\\
300D SPINN \cite{DBLP:journals/tacl/WilliamsDB18}&59.6&-\\
\midrule
(LSG) w/o NDD&32.5&25.7\\
(UTL) DIORA+PP&54.7&34.8\\
(UTL) DIORA+PP+POS&54.7&43.3\\
(LSG) Tight DP-NDD&59.3&\bf 55.4\\
(LSG) Loose DP-NDD&\bf 61.8&51.5\\
\bottomrule
\end{tabular}
}
\caption{Comparison on unlabeled and labeled F1 scores among methods for unsupervised constituency parsing on WSJ test dataset. \textbf{PP:} Post-processing heuristics. \textbf{Aug}: Rule-based Augmentation. *: Multiple edges are kept as constituents can have multiple labels.}
\label{tab:main}
\end{table}

Our main results and the comparison with previously reported results are shown in Table~\ref{tab:main}. We evaluate the models by unlabeled F1 for comparison with previous parsing methods. The performances of UTL and LSG are both reported to set baselines for those two frameworks. 

From Table~\ref{tab:main}, our DP-NDD-based LSG algorithm, DP-NDD with a loose configuration, outperforms all previous unsupervised methods for constituency parsing and remarkably boosts the state-of-the-art unlabeled F1 score to upper than $60.0$. Comparing with previous state-of-the-art methods consists of complex systems like post-processing with numerous linguistic rules, our algorithms are much simpler and of better scalability. We attribute this advance to the power of pre-trained language model which is able to cast constituents with high structural difference into near spaces in the latent space. 

For labeled F1 score, our algorithms also reach significant performance. DP-NDD with a tight configuration reaches a strong performance of $54.5$ which is even higher than most unlabeled F1 results from previous systems. Thus, we claim to successfully implement the first unsupervised full constituency parsing in recent years. Moreover, our method involves a much simpler PLM, BERT, than the highest baseline, xlnet, in \cite{DBLP:conf/iclr/KimCEL20}, but reaches a much higher performance ($13.5$ unlabeled F1 score). Comparing with result of BERT-base in \cite{DBLP:conf/iclr/KimCEL20}, unlabeled F1 score from DP-NDD is $19.5$, which shows the high efficiency of DP-NDD-based method. 

Comparing with UTL, LSG achieves higher F1 scores in both unlabeled and labeled treebanks. We conclude from this phenomenon that using label-specific method (Our molds are for a certain label) can extract constituents better than parsing spans of different labels with a unified algorithm like in other PLM-based method \cite{DBLP:conf/iclr/KimCEL20,DBLP:conf/acl/ShenTZBMC20}. For UTL, incorporating POS benefits labeling in this framework much as this lifts the unlabeled F1 score to $8.5$ higher.

We also launch an ablation study by removing DP-NDD scores from the LSG framework. LSG without DP-NDD returns all spans that satisfy the POS constraints. Without the guide of DP-NDD, the performance of LSG algorithm drops dramatically, even to half of the initial performance. We conclude from this phenomenon that our DP-NDD metric is essential for unsupervised full constituency parsing.

\section{Analysis and Discussion}

\subsection{Label-specific Evaluation}

\begin{table}
\centering
\small
\scalebox{0.99}{ 
\begin{tabular}{lccccc}
\toprule
Label & UR & LP & LR & LF1 & Prop.\\
\midrule
NP&66.20&68.49&64.34&66.35&42.08\%\\
VP&38.84&53.54&36.10&43.12&19.75\%\\
ADJP&51.20&14.97&28.89&19.72&2.02\%\\
ADVP&79.84&47.37&70.40&56.63&2.74\%\\
PP&63.84&66.37&51.53&58.02&12.40\%\\
\bottomrule
\end{tabular}
}
\caption{Performance of DP-NDD-based LSG algorithm on different labels. Unlabeled results (UR) are from loose DP-NPP and labeled (LP, LR, LF) results are from tight DP-NPP. \textbf{Prop.}: Proportion of labels in test treebanks.}
\label{tab:lsg}
\end{table}

We analyze the ability of our LSG algorithm for parsing edges of different labels in this section. We report unlabeled and labeled performance of LSG algorithm on different labels. Precision, recall and F1 score are all considered for labeled treebanks and only recall is evaluated for labeled treebanks. 

As presented in Table~\ref{tab:lsg}, LSG performs well on extracting noun, adverb and preposition phrases. For these phrases, LSG leads to high results in unlabeled recalls and labeled F1 scores. We mainly attribute the success of LSG to the high performance in discerning noun phrases, which take $42.08\%$ proportion of the constituents. LSG performs rather weaker for verb and adjective phrases as patterns of these phrases are more variable. Thereby, LSG will be more likely to confuse them with other phrases when trying to discern. This point will be elaborated in Section~\ref{5.3}. 

In contrast, phrases with regular patterns like adverb phrases and preposition phrases are more likely to be discerned successfully. This phenomenon can be attributed to the matching nature of our algorithm as less disturbance will be caused if a span is substituted by another span in a similar pattern. Take instances in Figure~\ref{fig:example} for explanation, substituting \textit{into the hole} with most preposition phrases will only result in subtle disturbance, i.e., \textit{in a warm autumn day}, \textit{before the crashing}. But verb phrase contain a variety of patterns like \textit{is so smart} and \textit{to enjoy their lunch}. Their substitution to the verb phrase \textit{jumps into the hole} will cause much more disturbance. Thus, the selection of molds for verb phrases should be more carefully to cover the patterns of verb phrases. But this still remains another problem that these patterns may be confused with other phrases like labeling \textit{to enjoy their lunch} to be a preposition phrase. Current structure-oriented LSG algorithm may not be able to offer a proper solution to this confusion, so we plan to leverage preciser semantics for a try in the future. 

\subsection{Labeling Performance}

\begin{table}
\centering
\small
\scalebox{0.99}{ 
\begin{tabular}{lcccccc}
\toprule
Label & P & R & F1 & P$^\dag$ & R$^\dag$ & F1$^\dag$\\
\midrule
NP&88.02&86.70&87.36&91.34&98.86&94.95\\
VP&99.17&50.70&67.10&98.52&90.26&94.21\\
ADJP&27.86&75.88&40.76&91.33&37.23&52.90\\
ADVP&63.19&55.74&59.23&93.93&85.15&89.32\\
PP&40.32&82.57&54.18&84.30&97.69&90.50\\
\bottomrule
\end{tabular}
}
\caption{Labeling performance of DP-NDD-based UTL algorithm on unlabeled golden edges in WSJ-10 treebanks. $\dag$: Refined by POS.}
\label{tab:utl}
\end{table}

We analyze the labeling performance of our UTL algorithm in this section. To avoid parsing bias caused by parser chosen for constructing unlabeled constituency trees, we follow \cite{DBLP:conf/naacl/DrozdovVYIM19}, we construct a WSJ-10 dataset by sampling sentences with length under 10 from train, development and test datasets. Then, constituents including noun, verb, adjective, adverb and preposition phrases are filtered from these sentences. WSJ-10 contains 17935 golden constituents and we use UTL algorithm to label constituents.

We report the experiment results of UTL in Table~\ref{tab:utl}. Without the refinement of POS information, UTL results in high precision and recall for labeling noun phrase, which verifies its capacity for discerning noun phrase's patterns. Adjective phrase remains to be the most difficult constituent for parsing and other phrases are of the medium parsing difficulty. POS-based refinement works for all phrases by significantly improving the F1 score of noun, verb, adverb and preposition phrases to around $90.0$ while still leaving the adjective phrase as a hard problem due to the difficulty in keeping recall and precision score for adjective phrase balanced. 

\subsection{Confusion in Constituent Discerning}
\label{5.3}

\begin{figure}
\centering
\includegraphics[width=0.50\textwidth]{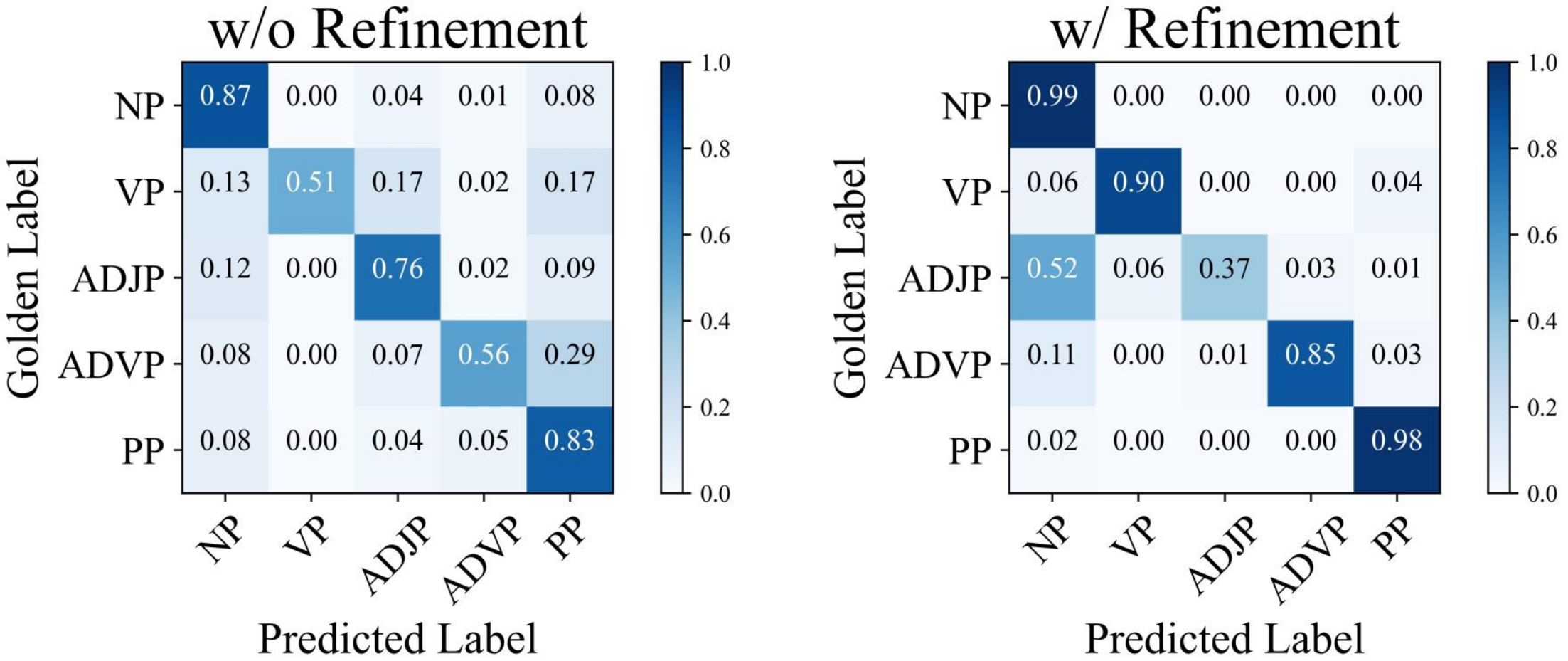}
\caption{Confusion matrix in labeling WSJ-10 dataset.}
\label{fig:cm}
\end{figure}

Following the discussion of labeling performance, we further analyze factors that affect the constituent discerning procedure. We depict the confusion matrix in Figure~\ref{fig:cm}. When POS is not used to help parsing, the most confusing labels are verb and adjective phrases. But the adjective phrase becomes prominently confusing when POS is taken into consideration, indicating that some adjective phrases have common POS patterns with noun phrases. 

To go deeper for the factors behind the confusion in labeling, we construct disturbance matrices by sampling constituent pairs from WSJ and WSJ-10 datasets. We sample $2000$ for each label pair and record the average POS-NDD caused by the substitution. Disturbance matrix is shown to be the direct reflection of pattern difference among constituents. Generally, self disturbance (disturbance between constituents of same labels) is lower than mutual disturbance (disturbance between constituents of different labels). Moreover, Phrases with more patterns like verb phrase have higher self disturbance. Referring to the confusion matrix without refinement, confusion appears when self disturbance is not enough lower than the mutual disturbance, i.e., VP-ADJP, VP-PP, ADVP-PP. The discerning difficulty leads to the drop in recall scores for verb and preposition phrases. For adjective phrases, its precision is affected to drop as some parts of noun phrases, which take a large proportion in constituents, are mislabeled as adjective phrases. 

\begin{figure}
\centering
\includegraphics[width=0.50\textwidth]{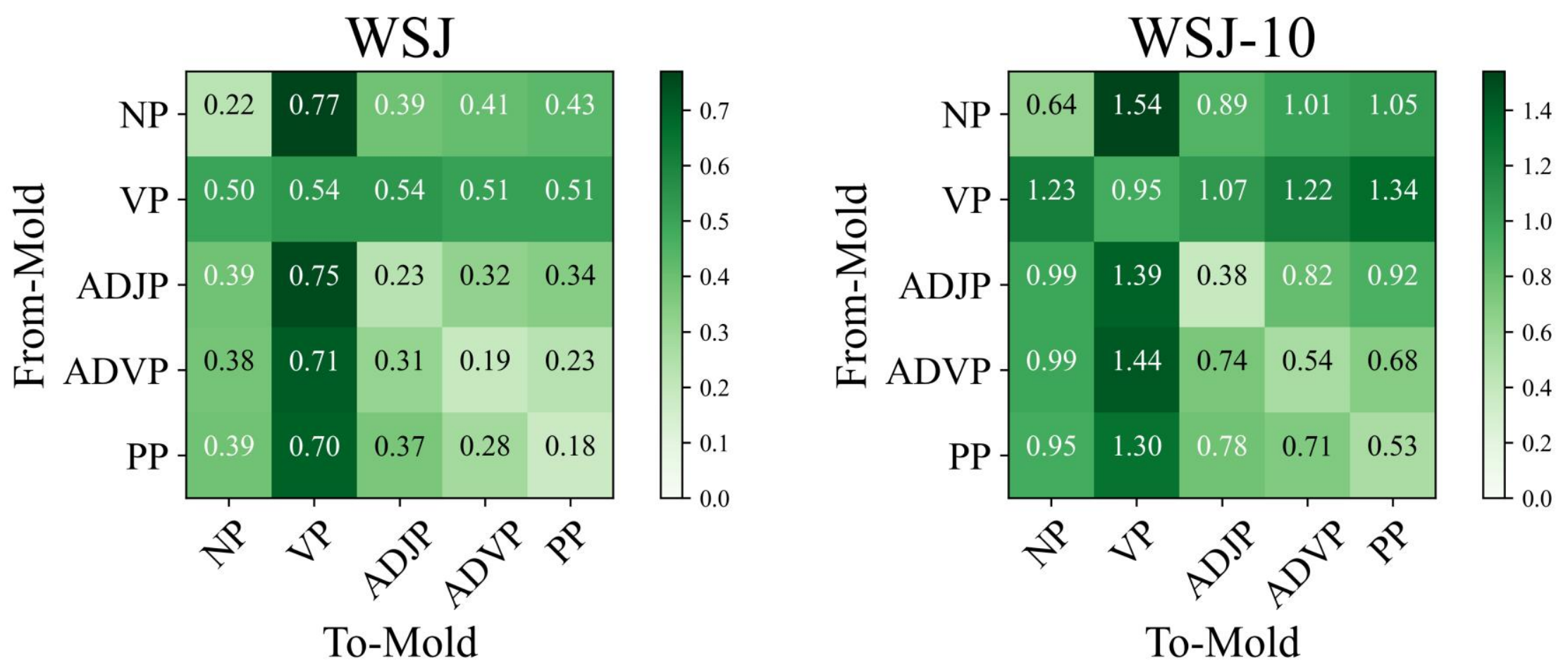}
\caption{Average semantic disturbance (POS-NDD) caused by constituent substitution.}
\label{fig:ndd_d}
\end{figure}

\subsection{How about other tasks?}

\begin{figure}
\centering
\includegraphics[width=0.50\textwidth]{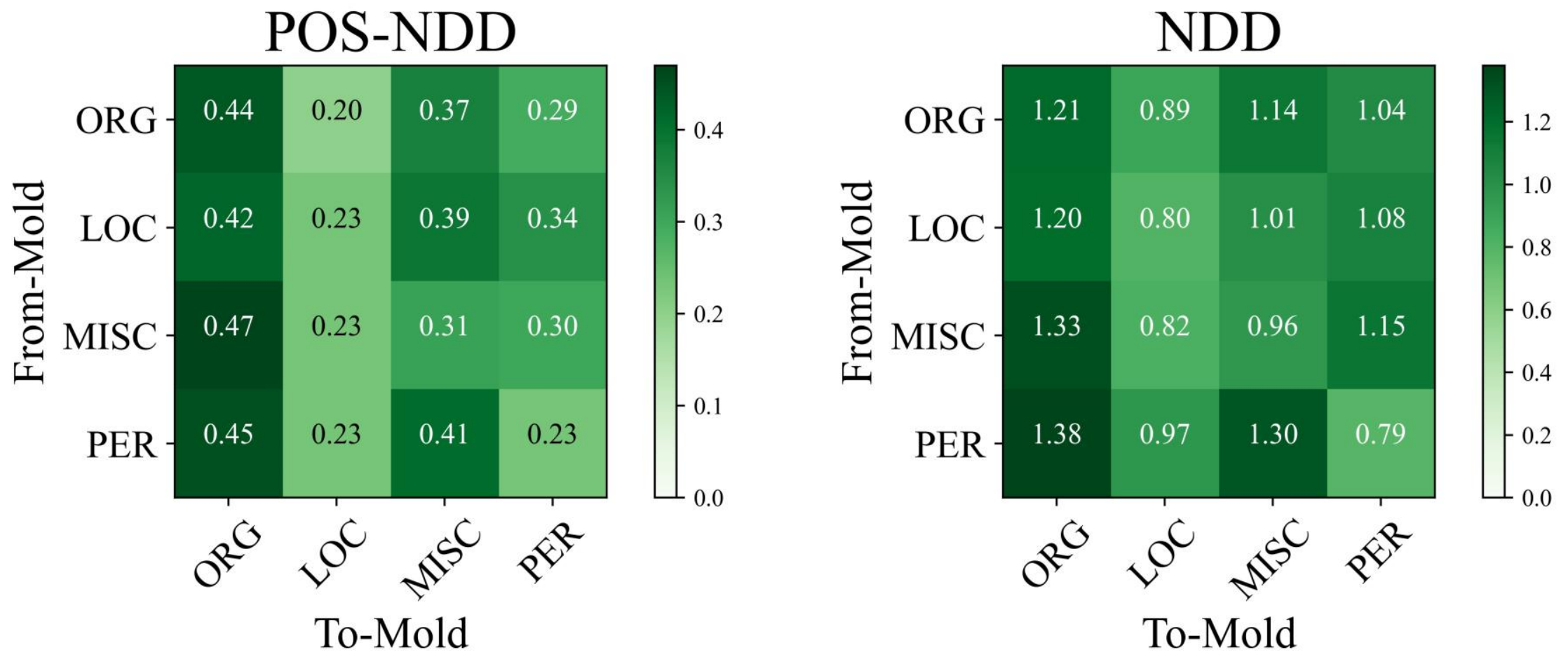}
\caption{Average semantic disturbance (POS \& POS-NDD) caused by constituent substitution for named entity extraction.}
\label{fig:ner}
\end{figure}

To verify the generalization of applying NDD for capturing labeled spans, we conduct experiments on named entity recognition, where all spans are noun phrases in constituency. We choose Conll-03 \cite{DBLP:conf/conll/SangM03} as the NER dataset. Conll-03 consists of named entities labeled in $4$ types: \textit{[ORG]}, \textit{[PER]}, \textit{[MISC]} and \textit{[PER]}. We sample 2000 pairs of spans in the same way that we do in~\ref{5.3}. We evaluate the average disturbance caused by substitute one span with another using both POS-NDD and the orginal NDD. 

Figure~\ref{fig:ner} shows the disturbance matrix for NER. Comparing with constituents, substitution using named entity on average results in much lower POS-NDD since named entities are all noun phrases as mentioned before. Generally, self disturbance is lower than mutual disturbance, making it a plausible to label named entities with NDD. Comparing with POS-NDD, NDD captures preciser semantics changes as described before. NER experiment results also support that NDD generally perform better in discerning named entity, which share structural similarity with each other, especially for disturbance caused by substitution to \textit{[LOC]}. 

Among labels, \textit{[PER]} is the easiest for discerning as it differs the most from other labels. In contrast, \textit{[ORG]} and \textit{[LOC]} are likely to be confused with each other as they play similar roles in semantics. For instance, we may say \textit{a meeting took place in \textbf{UN}} or \textit{a meeting took place in \textbf{Paris}}, but \textit{a meeting took place in \textbf{Jack}} is not semantically plausible. We conclude from the disturbance matrix the difficulty in entity labeling should be ranked as \textit{[ORG]}$>$\textit{[MISC]}$>$\textit{[LOC]}$>$\textit{[PER]}.


\section{Related Work}

\subsection{Unsupervised Constituency Parsing}

Since the introduction of language models pre-trained on large corpus like BERT \cite{DBLP:conf/naacl/DevlinCLT19}, extracting constituents from those models raises as a new way for unsupervised constituency parsing \cite{DBLP:conf/iclr/KimCEL20,DBLP:conf/acl/ShenTZBMC20}. These methods try to extract constituents by calculating the syntactic distance \cite{DBLP:conf/acl/BengioSCJLS18} which is supposed to reflect the information association among constituents according to \cite{DBLP:conf/iclr/ShenLHC18,DBLP:conf/emnlp/WangLC19}. The extraction of latent trees from PLMs has been studied on a variety of language models in \cite{DBLP:conf/iclr/KimCEL20}, which provides rich posterior knowledge for completing unsupervised constituency parsing.

Models trained on masked language model put forward another framework for unsupervised parsing procedure. These models, like DIORA and its variants \cite{DBLP:conf/naacl/DrozdovVYIM19,DBLP:conf/acl/SahayNMRI21}, have been verified by experiment results to be efficient in discerning constituents from sentences. Unfortunately, these models fail to label the constituents after constructing an unlabeled treebanks from sentences. Our method differs from previous work by using constituency molds to match constituents and thus induct their labels. Instead of figuring out direct relationships among words, we allow neighboring words to supervise the structural disturbance caused by substitution. As the result, our method enables labeling on the constituency tree which implements the full unsupervised constituency parsing. 

\subsection{Neighboring Distribution Divergence}

Neighboring distribution divergence \cite{DBLP:journals/corr/abs-2110-01176} is initially proposed to detect semantic changes caused by editions like compression \cite{DBLP:conf/emnlp/XuD19} or rewriting \cite{DBLP:conf/emnlp/LiuCLZZ20}. Their experiments on syntactic tree pruning and semantic predicate detection also show NDD to be aware of syntax and semantics. NDD is verified to have the capacity to detect predicates for semantic role labels by deleting or substituting words, which serves as our motivation to transfer this idea to unsupervised constituency parsing. We follow the idea in \cite{DBLP:journals/corr/abs-2110-01176} and further adapt it to extract and label constituents.

In previous years, there are other works which focus on leveraging pre-trained models to produce metrics reflecting syntactic or semantic information. To evaluate the quality of text generation, BERTScore \cite{DBLP:conf/iclr/ZhangKWWA20} matches representations from pre-trained language model of generated and golden sentences. Using pre-trained AMR parsers, \cite{DBLP:conf/eacl/OpitzF21} offers a explainable metric, MF-Score, for AMR-to-sentence generation. MF-Score assigns scores by reconstructing the AMR graphs to compare them with the golden ones and thus evaluates semantic similarity better than conventional sequence matching metrics like BLEU and ROUGE. Encouraged by our success in applying NDD for parsing, we plan to further explore these pre-trained model-based automatic metrics for more tasks. 

\section{Conclusion}
In this paper, we aim to explore an unsupervised full constituency parsing procedure which includes constituent labeling. We develop the recently proposed NDD metric into POS-NDD and exploit it by using dual mold to match constituents. Based on DP-NDD, we introduce two novel frameworks, labeled span generation and unlabeled tree labeling, which establish strong baselines for labeled constituency tree construction and set the new state-of-the-art for unlabeled F1 score. Further studies on constituents with NDD disclose the pattern variety of constituents with the same label and pattern similarity among constituents with different labels. Experiments on the NER dataset verify the generalization of our method to other tasks.

\bibliography{anthology,custom}

\begin{thebibliography}{26}
\expandafter\ifx\csname natexlab\endcsname\relax\def\natexlab#1{#1}\fi

\bibitem[{Chen et~al.(2015)Chen, Wang, and Zhao}]{chen-etal-2015-shallow}
Changge Chen, Peilu Wang, and Hai Zhao. 2015.
\newblock \href {https://doi.org/10.18653/v1/K15-2005} {Shallow discourse
  parsing using constituent parsing tree}.
\newblock In \emph{Proceedings of the Nineteenth Conference on Computational
  Natural Language Learning - Shared Task}, pages 37--41, Beijing, China.
  Association for Computational Linguistics.

\bibitem[{Choi et~al.(2018)Choi, Yoo, and Lee}]{DBLP:conf/aaai/ChoiYL18}
Jihun Choi, Kang~Min Yoo, and Sang{-}goo Lee. 2018.
\newblock \href
  {https://www.aaai.org/ocs/index.php/AAAI/AAAI18/paper/view/16682} {Learning
  to compose task-specific tree structures}.
\newblock In \emph{Proceedings of the Thirty-Second {AAAI} Conference on
  Artificial Intelligence, (AAAI-18), the 30th innovative Applications of
  Artificial Intelligence (IAAI-18), and the 8th {AAAI} Symposium on
  Educational Advances in Artificial Intelligence (EAAI-18), New Orleans,
  Louisiana, USA, February 2-7, 2018}, pages 5094--5101. {AAAI} Press.

\bibitem[{Devlin et~al.(2019)Devlin, Chang, Lee, and
  Toutanova}]{DBLP:conf/naacl/DevlinCLT19}
Jacob Devlin, Ming{-}Wei Chang, Kenton Lee, and Kristina Toutanova. 2019.
\newblock \href {https://doi.org/10.18653/v1/n19-1423} {{BERT:} pre-training of
  deep bidirectional transformers for language understanding}.
\newblock In \emph{Proceedings of the 2019 Conference of the North American
  Chapter of the Association for Computational Linguistics: Human Language
  Technologies, {NAACL-HLT} 2019, Minneapolis, MN, USA, June 2-7, 2019, Volume
  1 (Long and Short Papers)}, pages 4171--4186. Association for Computational
  Linguistics.

\bibitem[{Drozdov et~al.(2019)Drozdov, Verga, Yadav, Iyyer, and
  McCallum}]{DBLP:conf/naacl/DrozdovVYIM19}
Andrew Drozdov, Patrick Verga, Mohit Yadav, Mohit Iyyer, and Andrew McCallum.
  2019.
\newblock \href {https://doi.org/10.18653/v1/n19-1116} {Unsupervised latent
  tree induction with deep inside-outside recursive auto-encoders}.
\newblock In \emph{Proceedings of the 2019 Conference of the North American
  Chapter of the Association for Computational Linguistics: Human Language
  Technologies, {NAACL-HLT} 2019, Minneapolis, MN, USA, June 2-7, 2019, Volume
  1 (Long and Short Papers)}, pages 1129--1141. Association for Computational
  Linguistics.

\bibitem[{Kim et~al.(2020)Kim, Choi, Edmiston, and
  Lee}]{DBLP:conf/iclr/KimCEL20}
Taeuk Kim, Jihun Choi, Daniel Edmiston, and Sang{-}goo Lee. 2020.
\newblock \href {https://openreview.net/forum?id=H1xPR3NtPB} {Are pre-trained
  language models aware of phrases? simple but strong baselines for grammar
  induction}.
\newblock In \emph{8th International Conference on Learning Representations,
  {ICLR} 2020, Addis Ababa, Ethiopia, April 26-30, 2020}. OpenReview.net.

\bibitem[{Kim et~al.(2019)Kim, Dyer, and Rush}]{DBLP:conf/acl/KimDR19}
Yoon Kim, Chris Dyer, and Alexander~M. Rush. 2019.
\newblock \href {https://doi.org/10.18653/v1/p19-1228} {Compound probabilistic
  context-free grammars for grammar induction}.
\newblock In \emph{Proceedings of the 57th Conference of the Association for
  Computational Linguistics, {ACL} 2019, Florence, Italy, July 28- August 2,
  2019, Volume 1: Long Papers}, pages 2369--2385. Association for Computational
  Linguistics.

\bibitem[{Kitaev and Klein(2018)}]{DBLP:conf/acl/KleinK18}
Nikita Kitaev and Dan Klein. 2018.
\newblock \href {https://doi.org/10.18653/v1/P18-1249} {Constituency parsing
  with a self-attentive encoder}.
\newblock In \emph{Proceedings of the 56th Annual Meeting of the Association
  for Computational Linguistics, {ACL} 2018, Melbourne, Australia, July 15-20,
  2018, Volume 1: Long Papers}, pages 2676--2686. Association for Computational
  Linguistics.

\bibitem[{Lee et~al.(2013)Lee, Chang, Peirsman, Chambers, Surdeanu, and
  Jurafsky}]{DBLP:journals/coling/LeeCPCSJ13}
Heeyoung Lee, Angel~X. Chang, Yves Peirsman, Nathanael Chambers, Mihai
  Surdeanu, and Dan Jurafsky. 2013.
\newblock \href {https://doi.org/10.1162/COLI\_a\_00152} {Deterministic
  coreference resolution based on entity-centric, precision-ranked rules}.
\newblock \emph{Comput. Linguistics}, 39(4):885--916.

\bibitem[{Liu et~al.(2018)Liu, Zhu, and Shi}]{DBLP:conf/aaai/LiuZS18}
Lemao Liu, Muhua Zhu, and Shuming Shi. 2018.
\newblock \href
  {https://www.aaai.org/ocs/index.php/AAAI/AAAI18/paper/view/16347} {Improving
  sequence-to-sequence constituency parsing}.
\newblock In \emph{Proceedings of the Thirty-Second {AAAI} Conference on
  Artificial Intelligence, (AAAI-18), the 30th innovative Applications of
  Artificial Intelligence (IAAI-18), and the 8th {AAAI} Symposium on
  Educational Advances in Artificial Intelligence (EAAI-18), New Orleans,
  Louisiana, USA, February 2-7, 2018}, pages 4873--4880. {AAAI} Press.

\bibitem[{Liu et~al.(2020)Liu, Chen, Lou, Zhou, and
  Zhang}]{DBLP:conf/emnlp/LiuCLZZ20}
Qian Liu, Bei Chen, Jian{-}Guang Lou, Bin Zhou, and Dongmei Zhang. 2020.
\newblock \href {https://doi.org/10.18653/v1/2020.emnlp-main.227} {Incomplete
  utterance rewriting as semantic segmentation}.
\newblock In \emph{Proceedings of the 2020 Conference on Empirical Methods in
  Natural Language Processing, {EMNLP} 2020, Online, November 16-20, 2020},
  pages 2846--2857. Association for Computational Linguistics.

\bibitem[{Nguyen et~al.(2020)Nguyen, Nguyen, Joty, and
  Li}]{DBLP:conf/acl/NguyenNJL20}
Thanh{-}Tung Nguyen, Xuan{-}Phi Nguyen, Shafiq~R. Joty, and Xiaoli Li. 2020.
\newblock \href {https://doi.org/10.18653/v1/2020.acl-main.301} {Efficient
  constituency parsing by pointing}.
\newblock In \emph{Proceedings of the 58th Annual Meeting of the Association
  for Computational Linguistics, {ACL} 2020, Online, July 5-10, 2020}, pages
  3284--3294. Association for Computational Linguistics.

\bibitem[{Opitz and Frank(2021)}]{DBLP:conf/eacl/OpitzF21}
Juri Opitz and Anette Frank. 2021.
\newblock \href {https://aclanthology.org/2021.eacl-main.129/} {Towards a
  decomposable metric for explainable evaluation of text generation from
  {AMR}}.
\newblock In \emph{Proceedings of the 16th Conference of the European Chapter
  of the Association for Computational Linguistics: Main Volume, {EACL} 2021,
  Online, April 19 - 23, 2021}, pages 1504--1518. Association for Computational
  Linguistics.

\bibitem[{Peng et~al.(2021)Peng, Li, and
  Zhao}]{DBLP:journals/corr/abs-2110-01176}
Letian Peng, Zuchao Li, and Hai Zhao. 2021.
\newblock \href {http://arxiv.org/abs/2110.01176} {A novel metric for
  evaluating semantics preservation}.
\newblock \emph{CoRR}, abs/2110.01176.

\bibitem[{Sahay et~al.(2021)Sahay, Nasery, Maheshwari, Ramakrishnan, and
  Iyer}]{DBLP:conf/acl/SahayNMRI21}
Atul Sahay, Anshul Nasery, Ayush Maheshwari, Ganesh Ramakrishnan, and
  Rishabh~K. Iyer. 2021.
\newblock \href {https://doi.org/10.18653/v1/2021.findings-acl.436} {Rule
  augmented unsupervised constituency parsing}.
\newblock In \emph{Findings of the Association for Computational Linguistics:
  {ACL/IJCNLP} 2021, Online Event, August 1-6, 2021}, volume {ACL/IJCNLP} 2021
  of \emph{Findings of {ACL}}, pages 4923--4932. Association for Computational
  Linguistics.

\bibitem[{Sang and Meulder(2003)}]{DBLP:conf/conll/SangM03}
Erik F. Tjong~Kim Sang and Fien~De Meulder. 2003.
\newblock \href {https://aclanthology.org/W03-0419/} {Introduction to the
  conll-2003 shared task: Language-independent named entity recognition}.
\newblock In \emph{Proceedings of the Seventh Conference on Natural Language
  Learning, CoNLL 2003, Held in cooperation with {HLT-NAACL} 2003, Edmonton,
  Canada, May 31 - June 1, 2003}, pages 142--147. {ACL}.

\bibitem[{Shen et~al.(2018{\natexlab{a}})Shen, Lin, Huang, and
  Courville}]{DBLP:conf/iclr/ShenLHC18}
Yikang Shen, Zhouhan Lin, Chin{-}Wei Huang, and Aaron~C. Courville.
  2018{\natexlab{a}}.
\newblock \href {https://openreview.net/forum?id=rkgOLb-0W} {Neural language
  modeling by jointly learning syntax and lexicon}.
\newblock In \emph{6th International Conference on Learning Representations,
  {ICLR} 2018, Vancouver, BC, Canada, April 30 - May 3, 2018, Conference Track
  Proceedings}. OpenReview.net.

\bibitem[{Shen et~al.(2018{\natexlab{b}})Shen, Lin, Jacob, Sordoni, Courville,
  and Bengio}]{DBLP:conf/acl/BengioSCJLS18}
Yikang Shen, Zhouhan Lin, Athul~Paul Jacob, Alessandro Sordoni, Aaron~C.
  Courville, and Yoshua Bengio. 2018{\natexlab{b}}.
\newblock \href {https://doi.org/10.18653/v1/P18-1108} {Straight to the tree:
  Constituency parsing with neural syntactic distance}.
\newblock In \emph{Proceedings of the 56th Annual Meeting of the Association
  for Computational Linguistics, {ACL} 2018, Melbourne, Australia, July 15-20,
  2018, Volume 1: Long Papers}, pages 1171--1180. Association for Computational
  Linguistics.

\bibitem[{Shen et~al.(2019)Shen, Tan, Sordoni, and
  Courville}]{DBLP:conf/iclr/ShenTSC19}
Yikang Shen, Shawn Tan, Alessandro Sordoni, and Aaron~C. Courville. 2019.
\newblock \href {https://openreview.net/forum?id=B1l6qiR5F7} {Ordered neurons:
  Integrating tree structures into recurrent neural networks}.
\newblock In \emph{7th International Conference on Learning Representations,
  {ICLR} 2019, New Orleans, LA, USA, May 6-9, 2019}. OpenReview.net.

\bibitem[{Shen et~al.(2021)Shen, Tay, Zheng, Bahri, Metzler, and
  Courville}]{DBLP:conf/acl/ShenTZBMC20}
Yikang Shen, Yi~Tay, Che Zheng, Dara Bahri, Donald Metzler, and Aaron~C.
  Courville. 2021.
\newblock \href {https://doi.org/10.18653/v1/2021.acl-long.559} {Structformer:
  Joint unsupervised induction of dependency and constituency structure from
  masked language modeling}.
\newblock In \emph{Proceedings of the 59th Annual Meeting of the Association
  for Computational Linguistics and the 11th International Joint Conference on
  Natural Language Processing, {ACL/IJCNLP} 2021, (Volume 1: Long Papers),
  Virtual Event, August 1-6, 2021}, pages 7196--7209. Association for
  Computational Linguistics.

\bibitem[{Wang et~al.(2019)Wang, Lee, and Chen}]{DBLP:conf/emnlp/WangLC19}
Yau{-}Shian Wang, Hung{-}yi Lee, and Yun{-}Nung Chen. 2019.
\newblock \href {https://doi.org/10.18653/v1/D19-1098} {Tree transformer:
  Integrating tree structures into self-attention}.
\newblock In \emph{Proceedings of the 2019 Conference on Empirical Methods in
  Natural Language Processing and the 9th International Joint Conference on
  Natural Language Processing, {EMNLP-IJCNLP} 2019, Hong Kong, China, November
  3-7, 2019}, pages 1061--1070. Association for Computational Linguistics.

\bibitem[{Williams et~al.(2018)Williams, Drozdov, and
  Bowman}]{DBLP:journals/tacl/WilliamsDB18}
Adina Williams, Andrew Drozdov, and Samuel~R. Bowman. 2018.
\newblock \href {https://transacl.org/ojs/index.php/tacl/article/view/1281} {Do
  latent tree learning models identify meaningful structure in sentences?}
\newblock \emph{Trans. Assoc. Comput. Linguistics}, 6:253--267.

\bibitem[{Xu and Durrett(2019)}]{DBLP:conf/emnlp/XuD19}
Jiacheng Xu and Greg Durrett. 2019.
\newblock \href {https://doi.org/10.18653/v1/D19-1324} {Neural extractive text
  summarization with syntactic compression}.
\newblock In \emph{Proceedings of the 2019 Conference on Empirical Methods in
  Natural Language Processing and the 9th International Joint Conference on
  Natural Language Processing, {EMNLP-IJCNLP} 2019, Hong Kong, China, November
  3-7, 2019}, pages 3290--3301. Association for Computational Linguistics.

\bibitem[{Yogatama et~al.(2017)Yogatama, Blunsom, Dyer, Grefenstette, and
  Ling}]{DBLP:conf/iclr/YogatamaBDGL17}
Dani Yogatama, Phil Blunsom, Chris Dyer, Edward Grefenstette, and Wang Ling.
  2017.
\newblock \href {https://openreview.net/forum?id=Skvgqgqxe} {Learning to
  compose words into sentences with reinforcement learning}.
\newblock In \emph{5th International Conference on Learning Representations,
  {ICLR} 2017, Toulon, France, April 24-26, 2017, Conference Track
  Proceedings}. OpenReview.net.

\bibitem[{Zhang et~al.(2020{\natexlab{a}})Zhang, Kishore, Wu, Weinberger, and
  Artzi}]{DBLP:conf/iclr/ZhangKWWA20}
Tianyi Zhang, Varsha Kishore, Felix Wu, Kilian~Q. Weinberger, and Yoav Artzi.
  2020{\natexlab{a}}.
\newblock \href {https://openreview.net/forum?id=SkeHuCVFDr} {Bertscore:
  Evaluating text generation with {BERT}}.
\newblock In \emph{8th International Conference on Learning Representations,
  {ICLR} 2020, Addis Ababa, Ethiopia, April 26-30, 2020}. OpenReview.net.

\bibitem[{Zhang et~al.(2020{\natexlab{b}})Zhang, Zhou, and
  Li}]{DBLP:conf/ijcai/ZhangZL20}
Yu~Zhang, Houquan Zhou, and Zhenghua Li. 2020{\natexlab{b}}.
\newblock \href {https://doi.org/10.24963/ijcai.2020/560} {Fast and accurate
  neural {CRF} constituency parsing}.
\newblock In \emph{Proceedings of the Twenty-Ninth International Joint
  Conference on Artificial Intelligence, {IJCAI} 2020}, pages 4046--4053.
  ijcai.org.

\bibitem[{Zhong et~al.(2020)Zhong, Cambria, and
  Hussain}]{DBLP:journals/cogcom/ZhongCH20}
Xiaoshi Zhong, Erik Cambria, and Amir Hussain. 2020.
\newblock \href {https://doi.org/10.1007/s12559-020-09714-8} {Extracting time
  expressions and named entities with constituent-based tagging schemes}.
\newblock \emph{Cogn. Comput.}, 12(4):844--862.

\end{thebibliography}
\bibliographystyle{acl_natbib}

\newpage

\appendix
\section{Detailed Configuration}
\label{sec:config}

Before we release our codes, you can re-implement the results in our experiments with the configuration setting in this section. 

\subsection{Mold}

\begin{table}[H]
\centering
\small
\scalebox{0.83}{ 
\begin{tabular}{p{6.4cm}ccc}
\toprule
$W$ & $i$ & $j$ & $l$\\
\midrule
Influential members of the House Ways and Means Committee introduced legislation that would restrict how \underline{the new savings-and-loan} \underline{bailout agency} can raise capital , creating another potential obstacle to the government 's sale of sick thrifts . & 16 & 20 & NP$^\dag$\\
\midrule
\underline{The complex financing plan} in the S\&L bailout law includes raising \$ 30 billion from debt issued by the newly created RTC . & 1 & 4 & NP\\
\midrule
Another \$ 20 billion \underline{would be raised through} \underline{Treasury bonds , which pay lower interest rates} .&5&16&VP$^\dag$\\
\midrule
The bill \underline{intends to restrict the RTC to Treasury} \underline{borrowings only , unless the agency receives} \underline{specific congressional authorization} .&3&19&VP\\
\midrule
The complex financing plan in the S\&L bailout law includes raising \$ 30 billion from debt \underline{issued by the newly created RTC} .&17&22&VP\\
\midrule
But the RTC also requires `` working '' capital \underline{to maintain the bad assets of thrifts that are sold ,} \underline{until the assets can be sold separately} .&10&27&VP\\
\midrule
`` Such agency ` self-help ' borrowing is \underline{unauthorized and expensive} , far more expensive than direct Treasury borrowing , '' said Rep. Fortney Stark -LRB- D. , Calif. -RRB- , the bill 's chief sponsor .&9&11&ADJP$^\dag$\\
\midrule
`` Such agency ` self-help ' borrowing is unauthorized and expensive , \underline{far more expensive} than direct Treasury borrowing , '' said Rep. Fortney Stark -LRB- D. , Calif. -RRB- , the bill 's chief sponsor .&13&15&ADJP\\
\midrule
`` To maintain that dialogue is \underline{absolutely crucial} .&7&8&ADJP\\
\midrule
Many money managers and some traders had \underline{already} left their offices early Friday afternoon on a warm autumn day -- because the stock market was so quiet .&8&8&ADVP$^\dag$\\
\midrule
This country is \underline{fairly} big .&4&4&ADVP\\
\midrule
\underline{Therefore} , we can exchange in the market .&1&1&ADVP\\
\midrule
`` To maintain that dialogue is \underline{absolutely crucial} .&7&8&ADVP\\
\midrule
Once again -LCB- the specialists -RCB- were not able to handle the imbalances \underline{on the floor of the New York Stock Exchange} , '' said Christopher Pedersen , senior vice president at Twenty-First Securities Corp .&14&22&PP$^\dag$\\
\midrule
Big investment banks refused to step up to the plate to support the beleaguered floor traders \underline{by buying big blocks of stock} , traders say .&17&22&PP\\
\midrule
\underline{Just days after the 1987 crash} , major brokerage firms rushed out ads to calm investors .&1&6&PP\\
\bottomrule
\end{tabular}
}
\caption{Molds for result reproduction (from NP to PP). $\dag$: Used for UTL} 
\label{tab:mold(main)}
\end{table}

\begin{table}[H]
\centering
\small
\scalebox{0.83}{ 
\begin{tabular}{p{6.4cm}ccc}
\toprule
$W$ & $i$ & $j$ & $l$\\
\midrule
That debt would be paid off as the assets are sold , leaving the total spending for the bailout at \underline{\$ 50 billion} , or \$ 166 billion including interest over 10 years .&21&23&QP$^\dag$\\
\midrule
`` We would have to wait \underline{until we have collected} \underline{on those assets} before we can move forward , '' he said .&7&13&SBAR$^\dag$\\
\midrule
Instead , it settled on just urging the clients \underline{who are its lifeline} to keep that money in the market .&10&13&SBAR\\
\midrule
Influential members of the House Ways and Means Committee introduced legislation that would restrict how \underline{the new savings-and-loan} \underline{bailout agency can} \underline{raise capital} , creating another potential obstacle to the government 's sale of sick thrifts .&16&23&S$^\dag$\\
\midrule
Another \$ 20 billion would be raised through Treasury bonds , \underline{which} pay lower interest rates .&12&12&WHNP$^\dag$\\
\midrule
But the RTC also requires `` working '' capital to maintain the bad assets of thrifts \underline{that} are sold , until the assets can be sold separately .&16&17&WHNP\\
\midrule
Prices in Brussels , \underline{where} a computer breakdown disrupted trading , also tumbled .&5&5&WHADVP$^\dag$\\
\midrule
Dresdner Bank last month said it hoped to raise 1.2 billion marks \underline{-LRB- \$ 642.2 million -RRB-} by issuing four million shares at 300 marks each .&13&17&PRN$^\dag$\\
\midrule
Today 's Fidelity ad goes a step further , encouraging investors to stay \underline{in} the market or even to plunge in with Fidelity .&14&14&PRT$^\dag$\\
\bottomrule
\end{tabular}
}
\caption{Molds for result reproduction (the rest). $\dag$: Used for UTL} 
\label{tab:mold(ex)}
\end{table}

Table~\ref{tab:mold(main)} and~\ref{tab:mold(ex)} shows the molds we use for discerning constituents in LSG and labeling in UTL.

\newpage

\subsection{POS Constraint}

\begin{table}[H]
\centering
\small
\scalebox{0.66}{ 
\begin{tabular}{cp{2.0cm}p{2.0cm}p{2.0cm}p{2.0cm}c}
\toprule
Label& $\textrm{POS}(V[i])$ & $\textrm{POS}(V[j])$ & $\textrm{POS}(V[i-1])$ & $\textrm{POS}(V[j+1])$ & Max Len\\
\midrule
NP&NNP NNPS DT CD NN NNS JJ PRP PRP\$ \$&NN NNS NNP NNPS PRP CD POS&IN SOS , VB VBD CC VBZ VBG TO `` VBP VBN NN : RB NNS WRB '' RP CD&. IN , VBD VBZ VBP CC MD TO RB : NN JJ DT VBN NNP EOS WDT NNS VBG&-\\
\midrule
VP&VBD VB VBZ VBN VBP TO VBG MD&NN NNS NNP CD RB VB VBN JJ VBD PRP&NN NNS PRP TO NNP RB MD , VBZ VBD VBP WDT VB VBN CC IN '' DT&. , CC : NNP IN&-\\
\midrule
ADJP&JJ RB CD RBR JJR \$&JJ NN NNS CD JJR&DT VBZ VBD VBP VB IN RB VBN NN CC&. NN , IN NNS CC NNP&-\\
\midrule
ADVP&RB RBR&RB NN RBR IN&-&-&5\\
\midrule
PP&IN&NNP NNPS CD NN NNS PRP PRP\$&-&. , IN VBD CC : VBZ TO&-\\
\midrule
QP&\$ CD IN RB JJR&CD&IN VBD TO SOS VB VBZ DT :&IN NN NNS . , JJ TO DT&-\\
\midrule
SBAR&IN WDT PRP DT WRB WP TO&NN NNS NNP CD&NN , VBD NNS VBZ VBP SOS VB&. ,&-\\
\midrule
S&TO DT PRP VBG NNP NNS NN JJ VBD VBZ&NN NNS NNP VB RB CD JJ VBN&IN VBD WDT `` NN SOS , VBN CC VBZ WRB VB VBP WP&. ,&-\\
\midrule
WHNP&WDT WP WP\$&WDT WP&, NN NNS IN VBZ VB CC&VBD VBZ VBP MD DT PRP NNP RB NNS JJ IN ,&-\\
\midrule
WHADVP&WRB&WRB&-&-&-\\
\midrule
PRN&-LRB-&-RRB-&-&-&-\\
\midrule
PRT&RP&RP&-&-&1\\
\bottomrule
\end{tabular}
}
\caption{POS and length constraints for result reproduction. \textbf{SOS:} Start of the sentence. \textbf{EOS:} End of the sentence. -: No constraint.} 
\label{tab:constraint}
\end{table}

Table~\ref{tab:constraint} shows our constraints for POS and max length. These constraints are inducted by statistical property and simple linguistic rules.

\subsection{Threshold and Tolerance}

\begin{table}[H]
\centering
\small
\scalebox{0.9}{ 
\begin{tabular}{ccccc}
\toprule
Label&Threshold (t)& Tolerance (t)&Threshold (l)& Tolerance (l)\\
\midrule
NP&2.0&0.15&1.4&0.10\\
VP&0.8&0.15&2.0&0.05\\
ADJP&0.2&0.04&0.6&0.10\\
ADVP&0.8&0.03&0.8&0.03\\
PP&0.2&0.10&0.4&0.12\\
QP&0.2&0.03&0.2&0.03\\
SBAR&0.2&0.01&2.2&0.10\\
S&0.2&0.10&2.0&0.15\\
WHNP&1.0&0.10&1.0&0.10\\
WHADVP&1.0&0.10&1.0&0.10\\
PRN&1.0&0.10&1.0&0.10\\
PRT&1.0&0.10&1.0&0.10\\
\bottomrule
\end{tabular}
}
\caption{Thresholds and tolerances for result reproduction. \textbf{t}: Tight configuration. \textbf{l}: Loose configuration.} 
\label{tab:tt}
\end{table}


\end{document}